\documentclass[10pt,journal]{IEEEtran}
\ifCLASSOPTIONcompsoc
\else
\fi
\ifCLASSINFOpdf
  \usepackage[pdftex]{graphicx}
  \graphicspath{{../pdf/}{../jpeg/}}
  \DeclareGraphicsExtensions{.pdf,.jpeg,.png}
\else
  \usepackage[dvips]{graphicx}
  \graphicspath{{../eps/}}
  \DeclareGraphicsExtensions{.eps}
\fi
\usepackage{algorithm}
\usepackage{algpseudocode}
\usepackage{amsmath}
\usepackage{graphics}
\usepackage{epsfig}
\usepackage{times}
\usepackage{graphicx}
\usepackage{amssymb}
\usepackage{amsfonts}
\usepackage{color}
\usepackage{bm}
\usepackage{multirow}
\begin{document}
\title{Video Salient Object Detection via \\Fully Convolutional Networks}
%% author names and IEEE memberships
\author{Wenguan Wang, Jianbing Shen,~\IEEEmembership{Senior Member,~IEEE},
and Ling Shao,~\IEEEmembership{Senior Member,~IEEE}
%%%%%%%%%%%%%%%%%%%%%%%%%%%%%%%%%%%%%%%%%%%%%%%%%
\IEEEcompsocitemizethanks{%
\IEEEcompsocthanksitem
This work was supported in part by the National Basic Research Program of China (973 Program) (No. 2013CB328805),
the National Natural Science Foundation of China (No. 61272359), and the Fok Ying-Tong Education Foundation for Young Teachers.
Specialized Fund for Joint Building Program of Beijing Municipal Education Commission.
(Corresponding author: Jianbing Shen)
%%%%%%%%%%%%%%%%%%%%%%%%%%%%%%%%%%%%%%%%%%%%%%%%%%%%%%%%%%%
\IEEEcompsocthanksitem W. Wang and J. Shen are with Beijing Laboratory of Intelligent Information Technology,
School of Computer Science, Beijing Institute of Technology, Beijing 100081, P. R. China.
(email: shenjianbing@bit.edu.cn)
%%%%%%%%%%%%%%%%%%%%%%%%%%%%%%%%
\IEEEcompsocthanksitem L. Shao is with the School of Computing Sciences, University of East Anglia, Norwich NR4 7TJ, U.K.
(email: ling.shao@ieee.org )
%%%%%%%%%%%%%%%%%%%%%%%%%%%%%%%%
}% <-this % stops a space
\thanks{}
}
\markboth{IEEE Transactions on Image Processing} % ,~Vol.~6, No.~1, January~2011}%
{Shell \MakeLowercase{\textit{et al.}}: Bare Demo of IEEEtran.cls
for Computer Society Journals}

\maketitle
\begin{abstract}
This paper proposes a deep learning model to efficiently detect salient regions in videos. It addresses two important issues: (1) deep video saliency model training with the absence of sufficiently large and pixel-wise annotated video data; and (2) fast video saliency training and detection. The proposed deep video saliency network consists of two modules, for capturing the spatial and temporal saliency information, respectively.  The dynamic saliency model, explicitly incorporating saliency estimates from the static saliency model, directly produces spatiotemporal saliency inference without time-consuming optical flow computation. We further propose a novel data augmentation technique that simulates video training data from existing annotated image datasets, which enables
our network to learn diverse saliency information and prevents overfitting with the limited number of training videos. Leveraging our synthetic video data (150K video sequences) and real videos, our deep video saliency model successfully learns both spatial and temporal saliency cues, thus producing accurate spatiotemporal saliency estimate. We advance the state-of-the-art on the DAVIS dataset (MAE of .06) and the FBMS dataset (MAE of .07), and do so with much improved speed (2fps with all steps).
\end{abstract}

\begin{IEEEkeywords}
Video saliency, deep learning, synthetic video data, salient object detection, fully convolutional network.
\end{IEEEkeywords}

% \fi

\IEEEpeerreviewmaketitle

\section{Introduction}
\label{section1}
Saliency detection has recently attracted a great amount of research interest. The reason behind this growing popularity
lies in the effective use of these models in various vision tasks, such as image segmentation, object detection,
video summarization and compression, to name a few. Saliency models can be broadly classified into two categories:
human eye fixation prediction or salient object detection. According to the type of input, they can be further
categorized into static and dynamic saliency models. While static models take still images as input, dynamic models work on video sequences. In this paper, we focus on detecting distinctive regions in dynamic scenes.
\textit{Convolutional neural networks} (CNNs) have been successfully utilized in many fundamental areas of computer vision, including object detection \cite{girshick2015fast,ren2015faster}, semantic segmentation \cite{long2015fully}, and still saliency detection \cite{zhao2015saliency,li2016deep}. Inspired by this, we investigate CNNs to another computer vision task, namely video saliency detection.

The first problem of applying CNNs to video saliency is the lack of sufficiently large, densely labelled video training data. As far as we know, the successes of CNNs in computer vision are largely
attributed to the availability of large-scale annotated images (\textit{e.g.}, ImageNet \cite{russakovsky2015imagenet}).
However, existing video datasets are too small to provide adequate training data for CNNs. In Table 1, we list the statistics of the ImageNet dataset and widely adopted video object segmentation datasets, including FBMS \cite{brox2010object}, SegTrackV2 \cite{li2013video}, VSB100 \cite{galasso2013unified} and DAVIS \cite{Perazzi2016}. It can be observed that, the existing video datasets rarely match existing image datasets like ImageNet, in either quality or quantity. Besides, considering the high correlation between the frames from same video clip, existing video datasets are far unable to meet the needs of training CNNs for pixel-level video applications, like video salient object detection. On the other hand, for the moment, creating such a large-scale video dataset is usually infeasible, because annotating videos is complex and time-consuming. To this end, we propose a video data augmentation approach to synthetically generating labeled video training data, which explicitly leverages existing large-scale image segmentation datasets. The simulated video data are easily accessible and rapidly generated, close to realistic video sequences and present various motion patterns, deformations, companied with automatically generated annotations and optical flow. The experimental results via these automatically generated videos clearly demonstrate the practicability of our strategy.

%%%%%%%%%%%%%%%%%%%%%%%%%%%%%Table 1%%%%%%%%%%%%%%%%%%%%%%%%%%%%%%%%%%%%%%%%%%%%
\begin{table}%% [tbp]
\label{table1}
\caption{Statistics for ImageNet \cite{russakovsky2015imagenet}, FBMS \cite{brox2010object}, SegTrackV2 \cite{li2013video}, \protect\\VSB100 \cite{galasso2013unified} and DAVIS \cite{Perazzi2016} datasets.}
\centering
\begin{tabular}{c||c|c|c}  % {lccc}
\hline
Dataset &Ref &\#Clips &\#Annotations (frame/image)\\
\hline
ImageNet  &\cite{russakovsky2015imagenet}&- &$\thicksim 1.3\times10^6$\\
\hline
\hline
FBMS &\cite{brox2010object}&59  &$\thicksim$ 500~~\\
SegTrackV2 &\cite{li2013video}&14 &$\thicksim$ 1500\\
VSB100 &\cite{galasso2013unified}&100 &$\thicksim$ 600~~\\
DAVIS &\cite{Perazzi2016}&50 &$\thicksim$ 4000\\
\hline
\end{tabular}
\end{table}

Our video data synthesis approach clears the underlying challenge for learning CNNs for many applications in video processing, where dynamic saliency detection is of no exception. Another challenge for detecting saliency in dynamic scenarios derives from the natural demand of this task. As suggested by human visual perception research \cite{treisman1980feature,mital2013low}, when computing dynamic saliency maps, video saliency models need to consider both the spatial and the temporal characteristics of the scene. We propose a deep video saliency model for producing spatiotemporal saliency via fully exploring both the static and dynamic saliency information. The proposed model adopts fully convolutional networks (FCNs) \cite{long2015fully} for pixel-wise saliency prediction. Associated with existing rich image saliency data, the static saliency is deeply exploited and explicitly encoded in the deep learning process via transferring and fine-tuning recent success in image classification \cite{simonyan2014very}. For learning dynamic saliency cues, the proposed deep video saliency model learns from a large number of labelled videos, including both human-generated and natural video data, in a supervised learning mode. The static saliency is integrated into dynamic saliency detection process, thus for directly producing final spatiotemporal saliency estimation.

Another important contribution of this work is that our deep video saliency model is much more computationally
efficient compared with existing video saliency models.
Salient object detection is a key step in many image analysis tasks as it not only identifies relevant parts of a visual scene but may also reduce computational complexity by filtering out irrelevant segments of the scene. In recent years, some notable video saliency models have been proposed and show usefulness in many computer vision applications, such as video segmentation \cite{wang2015saliency} and video re-timing \cite{zhou2014time}. However, time efficiency becomes the common major bottleneck for the applicability of existing video saliency algorithms; most computation time has been spent for optical flow computation. Additionally, from the perspective of learning deep networks in dynamic scenes, many schemes \cite{simonyan2014two,bak2016two,khoreva2016learning} take optical flow as input, causing high computational expenses.

In this work, we propose a both effective and efficient video saliency model, which frees itself from the computationally expensive optical flow estimation. One of the key insights of this paper is that, unlike high-level video applications such as action detection, video saliency can derive from short-term analysis of video frames. Thus we directly capture temporal saliency via learning deep networks from frame pairs, instead of using long-term video information, such as optical flows from multiple adjacent video frames.

We comprehensively evaluate our method on the FBMS dataset \cite{brox2010object}, where the proposed video saliency model produces more accurate saliency maps than state-of-the-arts. Meanwhile, it achieves a frame rate of 2fps (including all steps) on a GPU. Thus it is a practical video saliency detection model in terms of both speed and accuracy. We also report results on the newly released DAVIS dataset \cite{Perazzi2016} and observe performance improvements over current competitors.

To summarize, the main contributions of this paper are threefold:
\begin{itemize}
\item We investigate convolutional neural networks for end-to-end training and pixel-wise saliency prediction in dynamic scenes. As far as we know, this is the first work for applying deep learning to video salient object detection.
\item We propose a novel training scheme based on synthetically generated video data, which explicitly leverages existing rich image datasets; both static and dynamic saliency information are encoded into a unified deep learning model.
\item Our methods are computationally efficient, much faster than traditional video saliency models and other deep networks in dynamic scenes.
\end{itemize}

The rest of this paper is structured as follows:
An overview of the related work is given in Section~\ref{section2}. Section~\ref{section3} defines our proposed deep saliency model.
The proposed synthetic video generation approach is articulated in
Section~\ref{section4}. Section~\ref{section5} shows experiment results on different
databases and compare with the state-of-the-art methods. Finally,
concluding remarks can be found in Section~\ref{section6}.

\section{Related work}
\label{section2}
In this section, we give a brief overview of recent works in two lines:
saliency detection, and deep learning models in dynamic scenes.

\subsection{Saliency Detection}
Saliency detection has been extensively studied in computer vision, and saliency models in general can be categorized into visual attention prediction or salient object detection. The former methods \cite{treisman1980feature,itti1998model,harel2006graph,judd2009learning} try to predict scene locations where a human observer may fixate. Salient object detection \cite{perazzi2012saliency,cheng2015global,wang2016correspondence} aims at uniformly highlighting the salient regions, which has been shown benefit to a wide range of computer vision applications. More detailed reviews of the saliency models can be found in \cite{borji2013state,borji2015salient}. Saliency models can be further divided into static and dynamic ones according to their input. In this work, we aim at detecting saliency object regions in videos.

Image saliency detection has been extensively studied for decades and most of the methods are driven by the well-known \textit{bottom-up} strategy. Early bottom-up models \cite{perazzi2012saliency,cheng2015global} are mainly based on detecting \textit{contrast}, assuming salient regions in the visual field would first pop out from their surroundings and computing feature-based contrast followed by various mathematical principles. Meanwhile, some other mechanisms \cite{wang2016correspondence,wei2012geodesic,zhu2014saliency} have been proposed to adopt some prior knowledge, such as \textit{background prior}, or global information, to detect salient objects in still images. More recently, deep learning techniques have been introduced to image saliency detection. These methods \cite{zhao2015saliency,li2015visual} typically use CNNs to examine a large number of region proposals, from which the salient objects are selected. Currently, more and more methods \cite{tang2016saliency,wang2016saliency,liu2016dhsnet,li2016deepsaliency} tend to learn in an end-to-end manner and directly generate pixel-wise saliency maps via fully convolutional networks (FCNs) \cite{long2015fully}.

Compared with saliency detection in still images, detecting saliency in videos is a much more challenging problem due to
the complication in the detection and utilization of temporal and motion information. So far, only a limited number of
algorithms have been proposed for spatiotemporal saliency detection. Early
models \cite{guo2008spatio,seo2009static,mahadevan2010spatiotemporal} can be viewed as simple extensions of exiting static saliency models with extra temporal dimension. Some more recent and notable approaches \cite{fang2014video,wang2015consistent,kim2015spatiotemporal,wang2015saliency,zhong2013video,Zhou2014} to this task have been proposed, showing inspired performance and good potentials in many computer vision applications \cite{Wang2017saliency,Wang2017super,Wang2017coseg,Wang2017cropping,wang2017stereoscopic}. However, the applicability of these approaches is severely limited by their high-computational costs. The main computational bottleneck comes from optical flow estimation, which contributes much to the promising results.

In recent years, the border of saliency detection has been extend to capturing common saliency among related images/videos \cite{Wang2015cosegment,wang2016higher,zhang2017co,zhang2016detection,zhang2016cosaliency}, inferring the salient event with video sequences \cite{Zhang2017} or scene understanding \cite{lu2017joint,yuan2015scene,nie2017}. However, there are significant differences between above methods and traditional saliency detection, especially considering their goals and core difficulties.

\subsection{Deep Learning Models in Dynamic Scenes}
In this section, we mainly focus on famous, deep learning models for computer vision applications in dynamic scenes, including action recognition \cite{simonyan2014two,charles2016personalizing}, object segmentation \cite{fragkiadaki2015learning,khoreva2016learning}, object tracking \cite{ma2015hierarchical,wang2015visual,wang2013learning,zhang2016robust,nam2015learning}, attention prediction \cite{bak2016two} and semantic segmentation \cite{tsai2016semantic}, and explore their architectures and training schemes. This will help to clarify how our approach differs from previous efforts and will help to highlight the important benefits in terms of effectiveness and efficiency.

Many approaches \cite{ma2015hierarchical,wang2015visual,tsai2016semantic} directly feed single video frames into neural networks trained on image data and adopt various techniques for post-processing the results with temporal or motion information. Unfortunately, these neural networks give up learning the temporal information which is often very important in video processing applications.

A famous architecture for training CNNs for action recognition in videos is proposed in \cite{simonyan2014two}, which incorporates two-stream convolutional networks for learning complementary information on appearance and motion. Other works \cite{bak2016two, fragkiadaki2015learning} adopt this architecture for dynamic attention prediction and video object segmentation . However, these methods train their models on multi-frame dense optical flow, which causes heavy computational burden.

In the areas of human pose estimation and video object processing, online learning strategy is introduced for improving performance \cite{khoreva2016learning,charles2016personalizing, wang2013learning,zhang2016robust,nam2015learning}. Before processing an input video, these approaches generate various training samples for fine-tuning the neural networks learned from image data, thus enabling the models to be optimized towards the object of interest in the test video sequence. Obviously, these models are quite time-consuming and the fine-tuned models are only specialized for specific classes of objects.

In this work, we show the possibilities of learning to detect generic salient objects in dynamic scenes by training on videos and images via an entirely offline manner. We proposed a novel technique for synthesizing video data via leveraging large amounts of image training data. The CNNs model can be efficiently and entirely trained on rich video sequences and images, thus successfully learning both static and dynamic saliency features. Meanwhile, it directly learns inner relationship between frames, getting rid of time-consuming motion computation. Thus, our algorithm is significantly faster than traditional video saliency methods and the deep learning architectures that demand optical flow as input. In summary, our CNNs model learns to detect video saliency in a fast and effective manner.
%%%%%%%%%%%%%%%%%%% Figure 1%%%%%%%%%%%%%%%%%%%%%%
\begin{figure}%%[t]
%%tr = 0.006, ts = 0.008
  \centering
      \includegraphics[width=\linewidth]{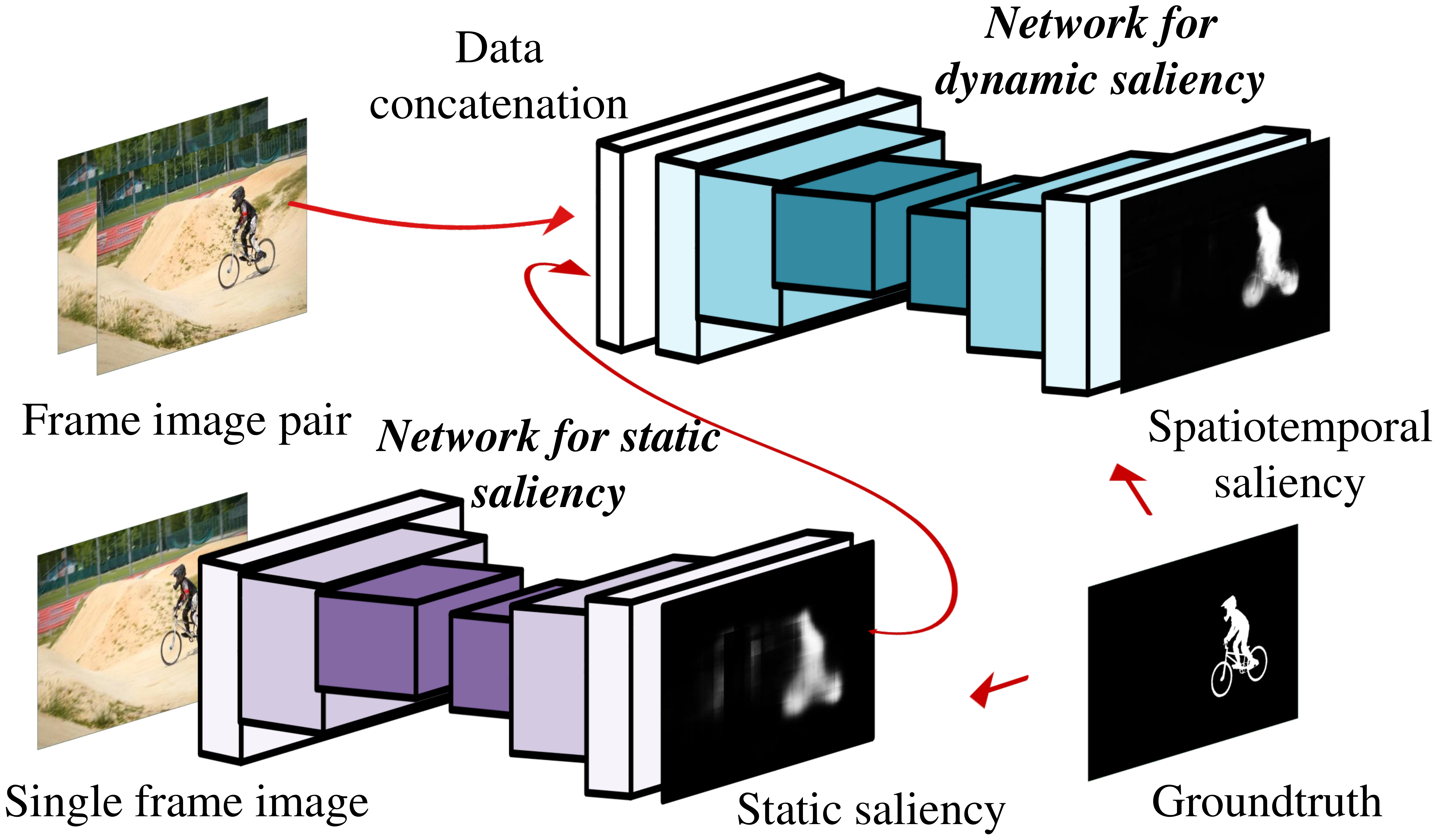}
\caption{A schematic representation of our proposed deep video saliency model. Our saliency model composes of two modules, which are designed for capturing the spatial and temporal saliency information simultaneously. The static saliency network (Sec.~\ref{section3.2}) takes single frame as input and outputs static saliency estimates. The dynamic saliency network (Sec.~\ref{section3.3}) learns dynamic saliency from frame pairs and takes static saliency generated by the first module as prior, thus producing the final spatiotemporal saliency maps. }
\label{fig1}
\end{figure}

\section{Deep Networks for Video Saliency Detection}
\label{section3}
In this work, we describe a procedure for constructing and learning deep video saliency networks using a novel synthetic video data generation approach. Our approach generates a large amount of video data (150K paired frames) from existing image datasets, and associates these annotated video sequences with existing video data to learn deep video saliency networks. We first introduce
the proposed CNNs based video saliency model in this section and then we describe our video synthesis approach in Sec.~\ref{section4}.

%Subsection text here.
\subsection{Architecture Overview}
\label{section3.1}
%%%%%%%%%%%%%%%%%%% Figure 2%%%%%%%%%%%%%%%%%%%%%%
\begin{figure*}%%[t]
%%tr = 0.006, ts = 0.008
  \centering
      \includegraphics[width=\linewidth]{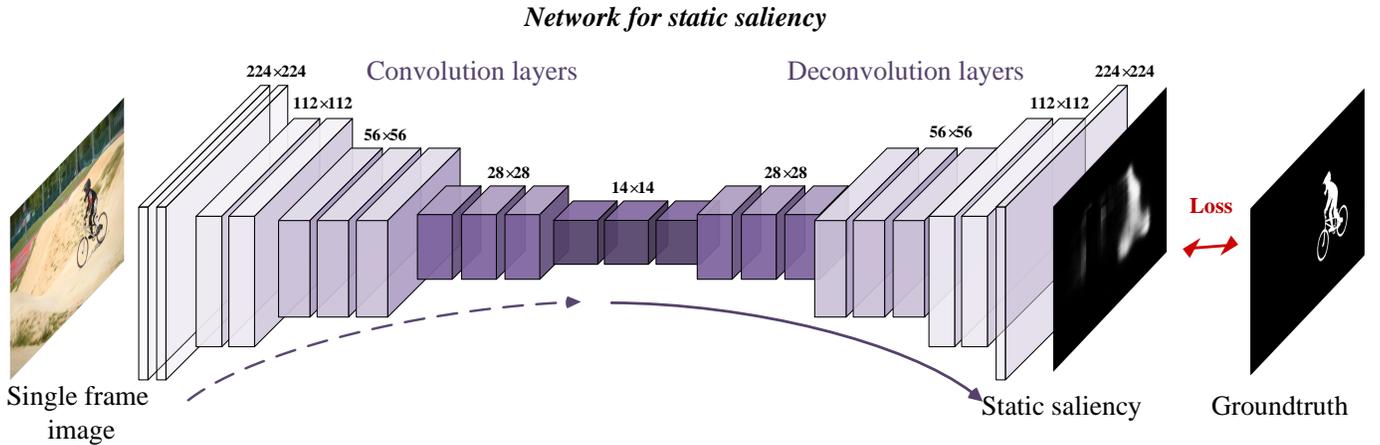}
\caption{Illustration of our network for static saliency detection. The network takes single frame image (for example, $224\times224$) as input, adopting multi-layer convolution networks that transforms the input image to multidimensional feature representation, then applying a stack of deconvolution networks for upsampling the feature extracted from the convolution networks. Finally, a fully convolution network with $1\times1$ kernel and \textit{sigmoid} activity function is used to output of a probability map in the same size as input, in which larger values mean higher saliency values.}
\label{fig2}
\end{figure*}
We start with an overview of our deep video saliency model before going into details below. At a high level, we feed frames of a video into a neural network, and the network successively outputs saliency maps where brighter pixels indicate higher saliency values. The network is trained with video sequences and images and learns spatiotemporal saliency in general dynamic scenes. Fig. \ref{fig1} shows the architecture of proposed deep video saliency model. Inspired by classical human visual perception research \cite{treisman1980feature,mital2013low}, which suggests both static and dynamic saliency cues contribute to video saliency, we design our model with two modules, simultaneously considering both the spatial and temporal characteristics of the scene.

The first module is for capturing static saliency, taking single frame image as input. It adopts fully convolutional networks (FCNs) for generating pixel-wise saliency estimate and utilizes previous excellent pre-trained models on large-scale image datasets. Boosted from rich image saliency benchmarks, this module is efficiently trained for capturing diverse static saliency information of interesting objects. This module is described in detail in Sec.~\ref{section3.2}.
The second module takes frame pairs and static saliency from the first module as input, and generates final dynamic saliency results. This network is trained from both synthetic and real labelled video data (see details in Sec.~\ref{section3.3}).

\subsection{Deep Networks for Static Saliency}
\label{section3.2}
A static saliency network takes a single frame image as input and produce a saliency map with the same size of the input. We model this process with a fully convolutional network (FCN). The bottom of this network is a stack of convolutional
layers. Convolutional layer is defined on shared parameters (weight vector and bias) architecture and has translation invariance characteristics. The input and output of each convolutional layer are a set of arrays, called feature maps, with size $h\times w \times c$, where $h$, $w$ and $c$ are height, width and the feature or channel dimensionality,
respectively. For the first convolutional layer, the input is the color image, with pixel size $h$ and $w$, and three channels. At the output, each feature map indicates a particular feature representation extracted at all locations on the input, which is obtained via convolving the input feature map with a trainable linear filter (or kernel) and adding a trainable bias parameter.
If we denote the input feature map as $X$, whose convolution filters are determined by the kernel weights $W$ and bias $b$, then the output feature map is obtained via:
\begin{equation}
    \begin{aligned}
    f_s(X;W,b) = W \ast_s X + b,
    \end{aligned}
    \label{eq:1}
\end{equation}
where $\ast_s$ is the convolution operation with stride $s$. After each convolutional layer, point-wise nonlinearity (\textit{e.g.}, ReLU) is applied for improving feature representation capability. Additionally, convolutional layers are often followed by some form of non-linear down-sampling (\textit{e.g.}, max pooling). This results in robust feature representation which tolerates small variations in the location of input feature map.

%%%%%%%%%%%%%%%%%%%% Figure 3%%%%%%%%%%%%%%%%%%%%%%
\begin{figure*}%%[t] %%tr = 0.006, ts = 0.008
  \centering
      \includegraphics[width=\linewidth]{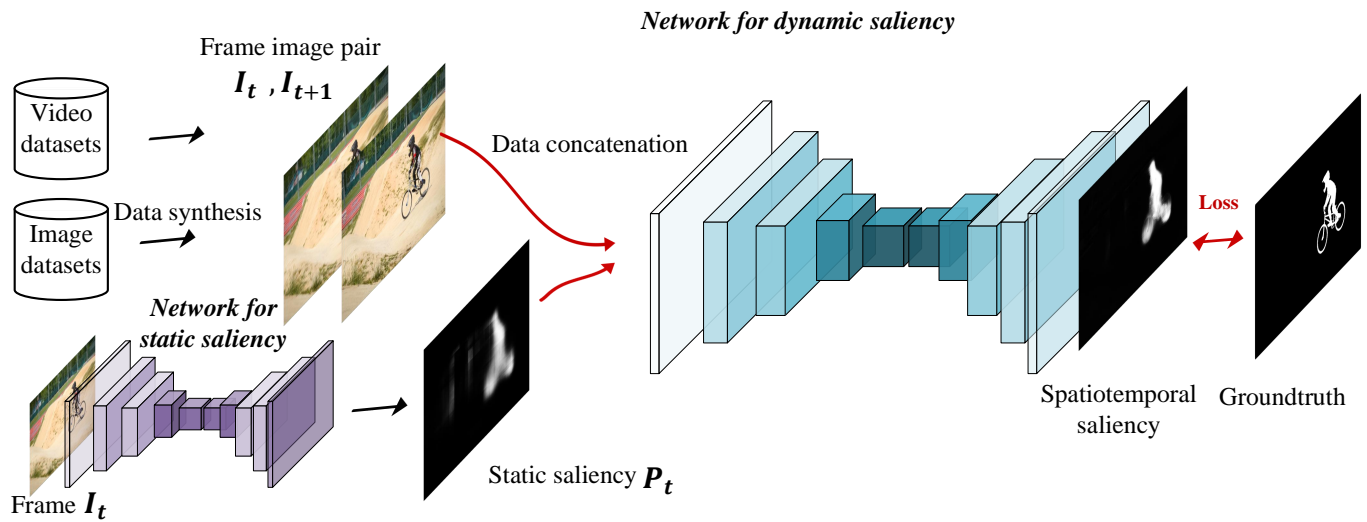}
\caption{Illustration of our network for dynamic saliency detection. Successive frame pairs $(I_t,I_{t+1})$ from real video data or synthesized from existing image datasets (described in Sec.~\ref{section4}), and static saliency information inferred from our static saliency network, are concatenated and fed into the dynamic network, which has a similar FCN architecture with the static network.  The dynamic network captures dynamic saliency, and considers static saliency simultaneously, thus directly generating spatiotemporal saliency estimation. }
\label{fig3}
\end{figure*}

Due to the stride of convolutional and feature pooling layers, the output feature maps are coarse and reduced-resolution. However, for saliency detection, we are more interested in pixel-wise saliency prediction. For upsampling the coarse feature map, multi-layer \textit{deconvolution} (or \textit{backwards convolution}) networks are put on the top of the convolution networks:
\begin{equation}
    \begin{aligned}
    Y = D_S(F_S(I;\Theta_{F});\Theta_{D}),
    \end{aligned}
    \label{eq:2}
\end{equation}
where $I$ is the input image; $F_S(\cdot)$ denotes the output feature map generated by the convolutional layers with total stride of $S$; $D_S(\cdot)$ denotes the deconvolution layers that upsample the input by a factor of $S$ to ensure the same spatial size of the output $Y$ and the input image $I$. The deconvolution operation is achieved via reversing the forward and backward passes of corresponding convolution layer.  All the parameters $\Theta$s of convolution and deconvolution layers are learnable.

Finally, on the top of the network, a convolutional layer with a $1\times1$ kernel is adopted for mapping the feature maps $Y$ into a precise saliency prediction map $P$ through a sigmoid activation unit. We use the sigmoid layer for pred so that each entry in the output has a real value in the range of 0 and 1. Due to the utilization of FCN, the network is allowed to operate on input images of arbitrary sizes, and preserves spatial information. Fig.~\ref{fig2} illustrates the detailed configuration of our deep network for static saliency. %For testing, the probability map actually is a salient map of the input image.

For training, all the parameters $\Theta$s are learned via minimizing a loss function, which is computed as the errors between the probability map and the ground truth. As demonstrated in \cite{mostajabi2015feedforward}, the use of an asymmetric weighted loss helps greatly in the case of unbalanced data. Considering the numbers of salient and non-salient pixels are usually imbalanced,
we compute a weighted cross-entropy loss. Given a training sample $(I, G)$ consisting
of an image $I$ with size $h\times w \times 3$, and groundtruth saliency map $G\in\{0, 1\}^{h\times w}$, the network produces saliency probability map $P \in [0, 1]^{h\times w}$. For any given training sample, the training
loss on network prediction $P$ is thus given by
\begin{equation}
    \begin{aligned}
    \mathcal{L}(P,G) = -\sum_{i=1}^{h\times w} \big((1-\alpha) g_i \log p_i + \alpha(1-g_i) \log (1-p_i)\big),
    \end{aligned}
    \label{eq:3}
\end{equation}
where $g_i\in G$ and $p_i\in P$; $\alpha$ refers to ratio of salient pixels in ground truth $G$.

We train the proposed architecture in an end-to-end manner. It is commonplace to initialize systems for many of vision tasks with a prefix of a network trained for image classification. This has shown to substantially reduce training time and improve accuracy. During training, our convolutional layers are initialized with the weights in the first five convolutional blocks of VGGNet \cite{simonyan2014very}, which was originally trained over 1.3 million images of the ImageNet dataset \cite{russakovsky2015imagenet}. The parameters of remaining layers are randomly initialized. Then we train our network with stochastic gradient descent (SGD) using backpropagation by minimizing the loss in Equ. \ref{eq:3}. More details of implementation are described in Sec.~\ref{section5.1}.

\subsection{Deep Networks for Dynamic Saliency}
\label{section3.3}
Now we describe our spatiotemporal saliency network.
As depicted in Fig. \ref{fig3}, the network has a similar structure as our static saliency network, which is based on FCN and includes multi-layer convolution and deconvolution nets. The dynamic network learns dynamic saliency information jointly with the static saliency results, thus directly generating spatiotemporal saliency estimates.

The training set consists of a collection of synthetic and real video data, which efficiently utilizes existing large-scale well-annotated image data (described in Sec. \ref{section4}). More specifically,
we feed successive pair of frames $(I_t,I_{t+1})$ and the groundtruth $G_t$ of frame $I_t$ in the training set into this network for capturing dynamic saliency. Meanwhile, since saliency in dynamic scenes is boosted by both static and dynamic saliency information, the network incorporates the saliency estimate $P_t$ generated by static saliency network as saliency priors indicative of potential salient regions. Thus our dynamic saliency network directly generates final spatiotemporal saliency estimates for frame $I_t$, which is achieved via exploring dynamic saliency cues and leveraging static saliency prior from the static saliency network.

We concatenate frame pair $(I_t,I_{t+1})$ and static saliency $P_t$ in the
channel direction, thus generating a tensor $\textbf{I}$ with size of $h\times w\times 7$. Then we feed $\textbf{I}$
into our FCN based dynamic saliency network, which has similar architecture of static saliency network. Only the first convolution layer is modified accordingly:
\begin{equation}
    \begin{aligned}
    f(\textbf{I};W,b) = W_{I_t} \ast I_t + W_{I_{t+1}} \ast I_{t+1} + W_{P_t} \ast P_t +b,
    \end{aligned}
    \label{eq:4}
\end{equation}
where $W$s represent corresponding convolution kernels; b is bias parameter. During training, stochastic gradient descent (SGD) is employed to minimize the weighted cross-entropy loss described before. After training, given a frame image pair and static saliency prior, the deep dynamic saliency model is able to output final spatiotemporal saliency estimate.
For testing, we first detect the static saliency map $P_{t}$ for frame $I_{t}$ via our static saliency network. Then frame image pair $(I_{t}, I_{t+1})$ and the static saliency map $P_{t}$ are fed into the dynamic saliency network for generating the final spatiotemporal saliency for frame $I_{t}$. After obtaining the video saliency estimate for frame $I_t$, we keep iterating this process for the next
frame $I_{k+1}$ until reaching the end of the video sequence. More implementation details can be found in Sec.~\ref{section5.1}. Qualitative and quantitative study of the effectiveness of our dynamic saliency model is described in Sec.~\ref{section5.3}.

Compared with the popular two-stream network structure used in \cite{simonyan2014two,fragkiadaki2015learning,bak2016two}, we merge the output of the static network into the dynamic saliency model, which directly produces spatiotemporal saliency results. This architecture brings two advantages. Firstly, the fusion of dynamic and static saliency is explicitly inserted into the dynamic saliency network, rather than training two-stream networks for spatial and temporal features and specially designing a fusion network for spatial and temporal feature integration. Secondly, the proposed model directly infers the temporal information from two adjacent frames instead of previous methods \cite{simonyan2014two,fragkiadaki2015learning} using optical flow images, thus our model gaining higher computation efficiency.
\section{Synthetic Video Data Generation}
\label{section4}
So far, we have described our networks for video saliency detection. We discuss our approach for training our networks for dynamic saliency below. As discussed in Sec.~\ref{section1}, existing video datasets \cite{brox2010object,li2013video,galasso2013unified,Perazzi2016} are insufficiently diverse and have very limited scales. As deep learning models are data-driven and have strong learning ability, directly learning deep networks on such video datasets would easily suffer overfitting. Noticing the gap between the requirement of learning neural networks for video processing and the lack of large-scale, high-quality annotated video data, we propose a technique for synthesizing video data from still frames.¡¡

Directly deriving video sequences from single image is also impossible. However, our video saliency network takes frame pairs as input, instead of the whole video sequence. That means we can simulates diverse but very short video sequences (only 2 frames in length) via fully utilizing well-labelled large-scale image datasets. Concretely, given a training sample $(I,G)$ from existing image saliency datasets, we wish to generate a pair of frames $(I,I')$, which
present various motion patterns, diverse deformations and smooth transformation, thus being close to real video signal. We start at simulating the correspondence between $I'$ and $I$, which is easier than directly inferring adjacent frame $I'$. Let $\textbf{x} = (x,y)$ denote a point position, the correspondence between $I'$ and $I$ can be represented as an optical flow field $\textbf{v} = (u,v)$ via:
\begin{equation}
    \begin{aligned}
    I'(\textbf{x}) = I(\textbf{x}+\textbf{v}(\textbf{x})).
    \end{aligned}
    \label{eq:5}
\end{equation}

The optical flow field $\textbf{v}$ directly represents the pixel-level motion information between two neighboring frames.
Next we only introduce how to set the vertical displacement $u$, as the method of generating $v$ is similar.

\begin{figure}%[h]
 \centering
   \includegraphics[width = \linewidth]{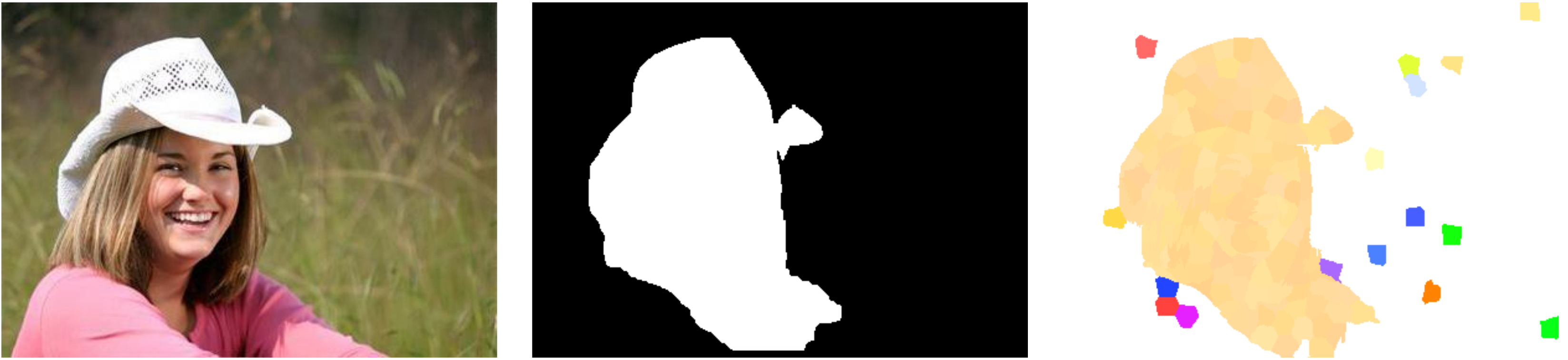}
  \\ \mbox{}\hfill \footnotesize{~~~~(a) Real frame $I$~~~~
     ~~(b) Saliency mask $G$~~
     (c) Initial optical flow $\textbf{v}$}
   \mbox{}\hfill
   \vspace*{3pt}
   \centering
   \mbox{}\hfill
   \includegraphics[width =\linewidth]{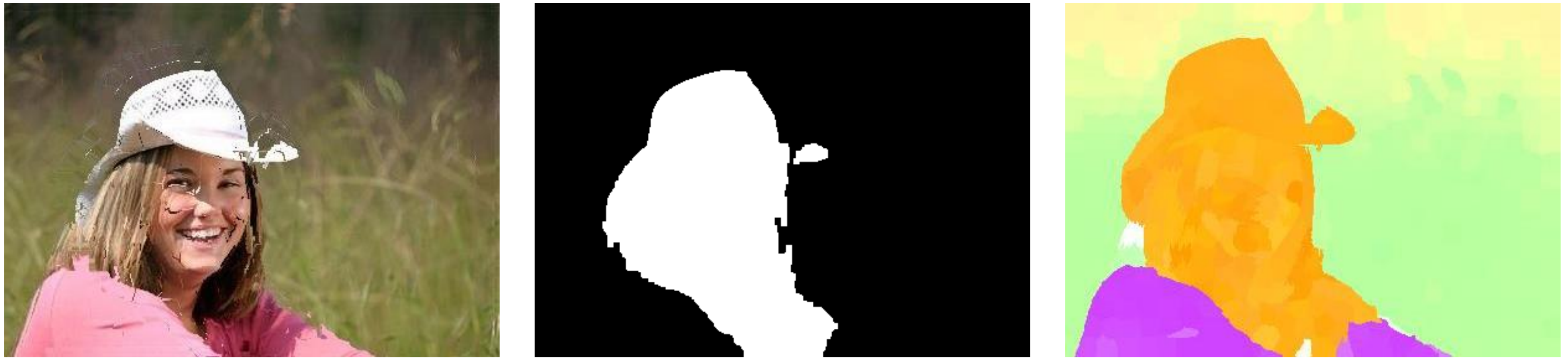}
  \\ (d) Synthetic frame $I'$ ~~
     (e) Saliency mask $G'$~~
     (f) Final optical flow $\bar{\textbf{v}}$
\caption{Illustration of our synthetic video data generation. A synthetic optical flow filed (c) is first initialized with considering various motion characters in real video sequences. Via Equ. \ref{eq:6}, final optical flow filed (f) is generated, which is more smooth and better simulates real motion patterns. According to (f), a synthetic frame image $I'$ and its saliency mask $G'$ are warped from (a) and (b), respectively.}
\label{fig:fig4}
\end{figure}

We model the optical flow on superpixel level as the motion of similar adjacent pixels should present \textit{consistency}. We oversegment $I$ into a group of superpixels $\mathcal{R}$. According to groundtruth label $G$, we further divide superpixels $\mathcal{R}$ into foreground superpixels $\mathcal{F}$ and background ones $\mathcal{B}$, where $\mathcal{R} = \mathcal{F}\bigcup \mathcal{B}$. For simulating the \textit{diverse motion patterns} of background, we randomly select $10\%$ background regions $\mathcal{S}$ from $\mathcal{B}$ and  randomly initialize their motion values $u$s (vertical displacement) from $[-d,d~\!]$, where $d = h/10$. The $u$s of the
other background regions are initialized as zero. The motion patterns of foreground are usually compactness, as the whole foreground regions move more regularly and purposefully compared with background. Beside, the motion between different foreground parts sometimes also present diverse. For example, the whole body of a person go an exact direction but his arms or legs may have different motions. For this, we first randomly set a value $m$ (from $[-d,d\!~]$) as the main motion patterns of the foreground regions. Then we randomly set $v$s of foreground regions from $[m-d/10,m+d/10]$ for representing the difference between foreground regions. This initialization process is visualized in Fig. \ref{fig:fig4}-a.

A similar process is adopted for generating the initial horizontal motion displacement ($v$) and we are able to get an initial optical flow $\textbf{v}$ for $I$. Next, we propose an energy function for smoothing and propagating the initial optical flow globally, yet preserving the difference between foreground and background in motion patterns. Let the initial motion vector of each superpixel $r_i\in \mathcal{R}$ be denoted as $\textbf{v}_i$, the final
motion vector $\bar{\textbf{v}}_i$ is obtained via optimizing the energy function as follows\footnote{Here we slightly reuse $\textbf{v}$ for representing the optical flow vector of superpixel without ambiguity.}:
 \begin{equation}
    \begin{aligned}
    \mathcal{E}(\bar{\textbf{v}},\textbf{v}) = \underbrace{\sum_{i}\lambda_i(\bar{\textbf{v}}_i-\textbf{v}_i)
    ^{2}}_{Unary~Term} + \underbrace{\sum_{i,i'\in \aleph}w_{i,i^{'}}(\bar{\textbf{v}}_i-\bar{\textbf{v}}_{i'})^{2}}_{Smooth~Term}.
    \end{aligned}
    \label{eq:6}
 \end{equation}
The first term is the unary constraint that each superpixel tends to have its initial motion, while the smooth term gives the interactive constraint that neighboring
superpixels have consistent motion patterns when their representative colors are similar. The superpixel neighborhood
set $\aleph$ contains all the spatially adjacent superpixels\footnote{For further encouraging the motion consistency of background regions, we consider all the selected background regions $\mathcal{S}$ are adjacent in neighboring system $\aleph$.}.
The parameter $\lambda$ is a positive coefficient measuring how much we want to fit the initial
motion. Typically, $\lambda = +\infty$ imposes the hard constraint that each region definitely has the initial motion. We define $\lambda$:
\begin{equation}
    \begin{aligned}
    \lambda_i = \left\{
        \begin{aligned}
            &1 ~~~~~~~~~~~~~~~~~~~~~~~~\text{if } r_i\in \mathcal{F} \\
            &1 ~~~~~~~~~~~~~~~~~~~~~~~~\text{if } r_i\in \mathcal{S} \\
            &10^{-4} ~~~~~~~~~~~~~~~~~~~~\text{otherwise }
        \end{aligned}
    \right.
    \end{aligned}
    \label{eq:7}
\end{equation}
For the seed regions (selected background regions $\mathcal{S}$ and all the foreground regions $\mathcal{F}$), we expect that they tend to preserve their initial motions; however, for other regions ($\mathcal{B}\setminus\mathcal{S}$), we emphasize more influence on the smooth term thus we can propagate the initial motions from those seed regions.

The weighting function $w_{i,i^{'}}$ in Equ. \ref{eq:6} defines a similarity measure for adjacent superpixels ($r_i,r_{i'}\in \aleph$):
\begin{equation}
    \begin{aligned}
    w_{i,i^{'}} = \left\{
        \begin{aligned}
            &exp^{-\parallel C(r_i)- C(r_{i'})\parallel^2} ~~~~~~~\text{if } r_i,r_{i'}\in \mathcal{F} \\
            &exp^{-\parallel C(r_i)- C(r_{i'})\parallel^2} ~~~~~~~\text{if } r_i,r_{i'}\in \mathcal{B} \\
            &0 ~~~~~~~~~~~~~~~~~~~~~~~~~~~~~\text{otherwise }
        \end{aligned}
    \right.
    \end{aligned}
    \label{eq:8}
\end{equation}
where $C(r)$ indicates the mean color vector of pixels in superpixel $r$. We set the weight $w_{i,i^{'}}$ as zero, when two adjacent superpixels are from foreground $\mathcal{F}$ and background $\mathcal{B}$, respectively. We consider motion consistency inside the foreground and background, while preserve motion difference between foreground and background.
Equ. \ref{eq:6} can be efficiently solved by convex optimization and we can obtain a smooth optical flow field $\textbf{v}$. As shown in Fig. \ref{fig:fig4}, base on $\textbf{v}$, we can generate a simulated frame $I'$ and its corresponding annotation $G'$ from $(I,G)$.

%%%%%%%%%%%%%%%%%%%% Figure 5%%%%%%%%%%%%%%%%%%%%%%
\begin{figure}%%[t] %%tr = 0.006, ts = 0.008
  \centering
      \includegraphics[width=0.99 \linewidth]{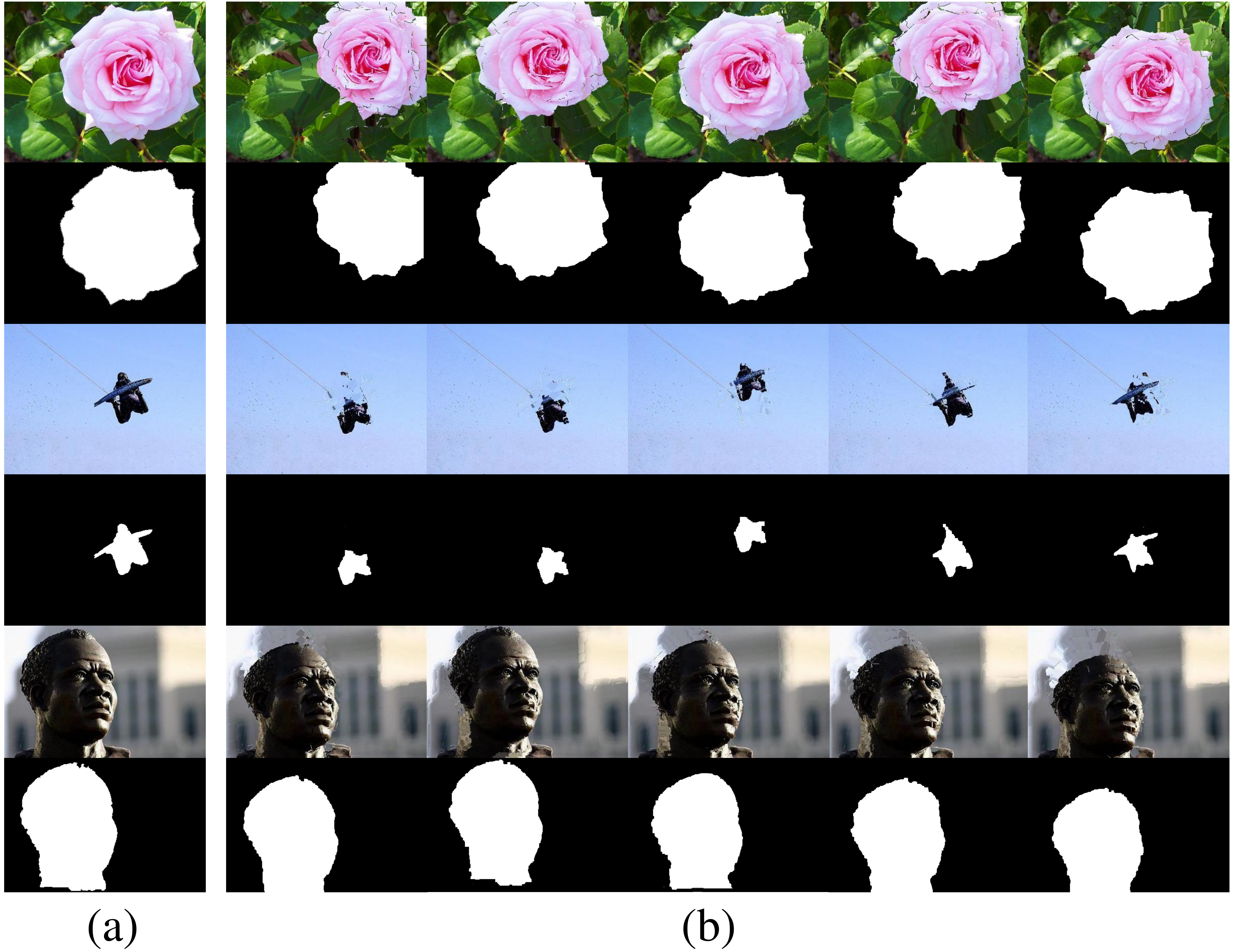}
\caption{(a) Real images and corresponding saliency groundtruth masks from existing image datasets. (b) Synthetic image examples and saliency masks generated via our method.}
\label{fig:fig5}
\end{figure}
The proposed method is very fast and outputs synthesized video frame pair, optical flow, and pixel-wise annotations simultaneously. The number of samples in existing image segmentation/saliency datasets is ten or hundred order of magnitude larger than in the video segmentation datasets, allowing us to generate enough scenes. For each
image sample $I$ of an image dataset, we generate ten simulated frames. Some simulated results can be observed in Fig. \ref{fig:fig5}. In our experiments, we use two large image saliency datasets MSRA10K~\cite{Liu2007} and DUT-OMRON~\cite{yang2013saliency}, generating more than $150K$ simulated videos associated with pixel-level annotations and optical flow within 3 hours (processing speed of 14 fps on one CPU). Those synthesized video data, combined with real video samples from existing video segmentation datasets, are fed into our model for learning general dynamic saliency information without over-fitting.

%%%%%%%%%%%%%%%%%%%%%%%%%%%%%%%%%%%%%
\section{Experimental Results}
\label{section5}
In this section, we describe our evaluation protocol and implementation details
(Sec.~\ref{section5.1}), provide exhaustive comparison results over two large datasets (80
videos in total, Sec.~\ref{section5.2}), study the quantitative importance of the different
components of our system (Sec.~\ref{section5.3}), and assess its
computational load (Sec.~\ref{section5.4}).

\subsection{Experimental Setup}
\label{section5.1}
\subsubsection{Datasets}
We report our performance on two public benchmark datasets:
Freiburg-Berkeley Motion Segmentation (FBMS) dataset \cite{brox2010object}, and Densely Annotated VIdeo Segmentation (DAVIS) dataset \cite{Perazzi2016}.
The FBMS dataset contains 59 natural video sequences, covering various challenges such as
large foreground and background appearance variation, significant
shape deformation, and large camera motion. This dataset is originally used for motion segmentation, where unsalient but moving objects are also labeled as foreground. We offer more precise annotations for this dataset via only labeling the main salient objects. The FBMS dataset comes with a split into a training set and a test set, where the training set includes 29 video sequences and the test set has 30 video sequences.
We also report our performance on the newly developed DAVIS dataset, which is one of the most challenging video segmentation benchmarks. It consists of 50 video sequences in total, and fully-annotated pixel-level segmentation ground-truth for each frame
is available. We report the performance of our method and other alternatives on the \textit{test set} of
FBMS dataset and the \textit{whole} DAVIS dataset.

%%%%%%%%%%%%%%%%%%%% Figure 6%%%%%%%%%%%%%%%%%%%%%%
\begin{figure*}%%[t] %%tr = 0.006, ts = 0.008
  \centering
      \includegraphics[width=\linewidth]{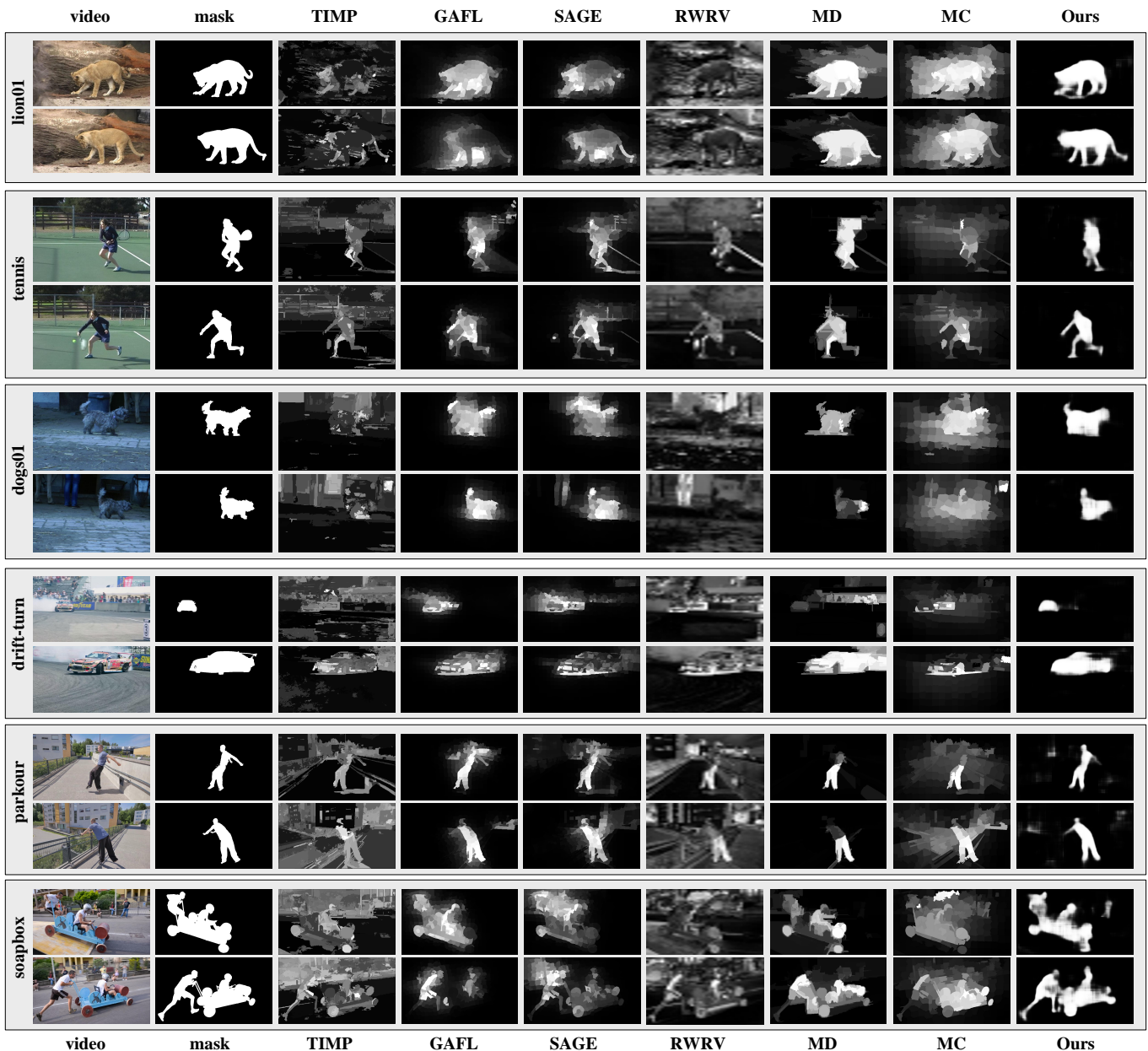}
\caption{Qualitative comparison against the state-of-the-art methods on the FBMS dataset videos \cite{brox2010object}  (\textit{lion01}, \textit{tennis} and \textit{dogs01}), and DAVIS dataset sequences \cite{Perazzi2016} (\textit{drift-turn}, \textit{parkour} and \textit{soapbox}) with pixel-level ground-truth labels. Our saliency method renders the entire objects as salient in complex scenarios, yielding continuous saliency maps that are most similar  to the ground-truth.}
\label{fig:fig6}
\end{figure*}

For training, we use two large image saliency datasets: MSRA10K~\cite{Liu2007} and DUT-OMRON~\cite{yang2013saliency}. The MSRA10K dataset comprising of $10K$ images, is widely used for saliency detection and covers a large variety of image contents -- natural
scenes, animals, indoor, outdoor, etc. Most of the images have a single salient object. The DUT-OMRON dataset is one of the most challenging image saliency datasets and contains 5172 images with multiple objects with complex structures and high background clutter. All the above datasets contain
manually annotated groundtruth saliency. The video sequences of the whole SegTrackV2 dataset \cite{li2013video} and the training set of the FBMS dataset are also used for training the dynamic saliency network, which include about 3K frame pairs\footnote{Due to the number of annotations provided by FBMS is very limited (only 4$\thicksim$6 frames are labeled for each video sequence), we provide extra $\thicksim$500 annotations.}.

\subsubsection{Implementation}
The proposed deep video saliency network has been implemented with the popular Caffe library \cite{jia2014caffe}, an open source framework for CNNs training and testing. For our static video saliency network, the weights
of the first five convolutional blocks are initialized by the VGGNet model \cite{simonyan2014very} trained on ImageNet \cite{russakovsky2015imagenet}, the other convolutional layers are initialized from zero mean Gaussian with a standard deviation of 0.01 and the biases are set to 0.  Based on this, our network was trained on the MSRA10K~\cite{Liu2007} and the DUT-OMRON~\cite{yang2013saliency} datasets with $100K$ iterations for saliency detection in static scenes.
Our dynamic video saliency network is also initialized from the VGGNet network. For the first convolutional layer, we use Gaussian initialization due to a different input channel from VGGNet. Benefiting from our video data synthesis approach, we can employ images and
annotations from existing saliency segmentation datasets for training our video saliency model.  The images and masks from  MSRA10K~ and DUT-OMRON~ datasets are used to generate more than $150K$ video slits. Then we combine our simulated video data with real video data ($\thicksim$3K frame pairs) from exiting video segmentation datasets \cite{li2013video,brox2010object} for generating an aggregate video saliency training set.
Our whole video saliency model is trained for $300K$ iterations.

For both two networks, we use stochastic gradient descent (SGD) and a polynomial learning policy with initial learning rate of $10^{-7}$. The momentum and weight decay are set to 0.9 and 0.0005. The whole training process costs about 40 hours on a PC with 3.4 GHz CPU, a TITANX GPU, and 32G RAM.
%%%%%%%%%%%%%%%%%%%% Figure 7%%%%%%%%%%%%%%%%%%%%%%
\begin{figure*}%%[t]
 %%tr = 0.006, ts = 0.008
  \centering
     \includegraphics[width=\linewidth]{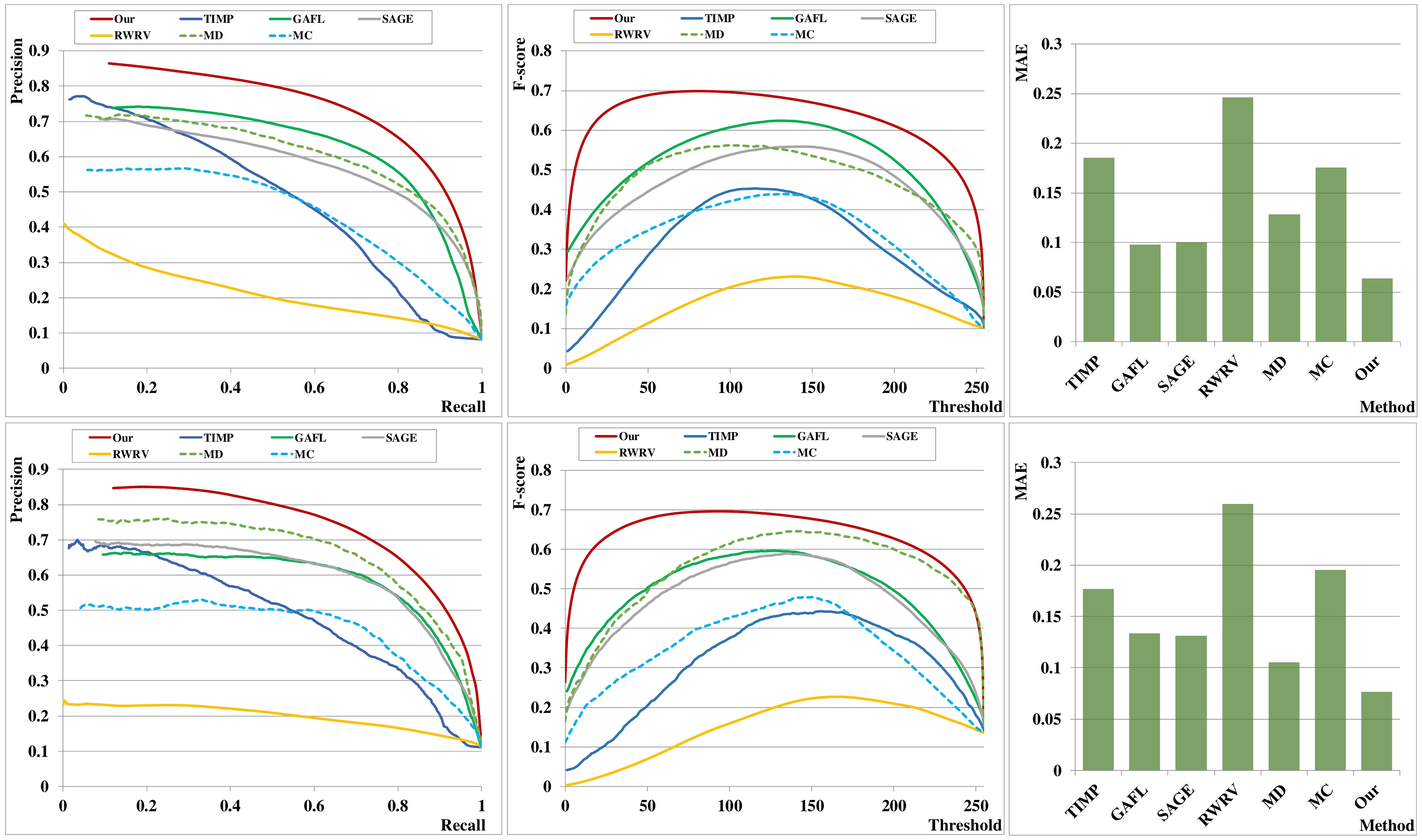} \\
     \mbox{}\hfill (a) \hfill\mbox{}
     \mbox{}\hfill (b) \hfill\mbox{}
     \mbox{}\hfill (c) \hfill\mbox{}
\caption{Comparison with 8 alternative saliency detection methods using the DAVIS dataset \cite{Perazzi2016} (\textbf{top}), and the  test set of the FBMS dataset \cite{brox2010object} (\textbf{bottom}) with pixel-level ground-truth: (a) average precision recall curve by segmenting saliency maps using fixed thresholds, (b) F-score, (c) average MAE. Notice that, our algorithm consistently outperforms other methods across different metrics.}
 \label{fig:fig7}
\end{figure*}
\subsection{Performance Comparison}
\label{section5.2}
To evaluate the quality of the proposed approach, we provide in this section quantitative comparison for performance of the proposed method against various top-performing alternatives: saliency via deep feature (MD) \cite{li2015visual}, saliency via absorbing markov chain (MC) \cite{jiang2013saliency}, space-time saliency for time-mapping (TIMP) \cite{zhou2014time}, gradient-flow filed based saliency (GAFL) \cite{wang2015consistent}, geodesic distance based video saliency (SAGE) \cite{wang2015saliency}, and saliency via random walk with restart (RWRV) \cite{kim2015spatiotemporal}, on test set (30 video sequences) of the FBMS dataset and the whole DAVIS dataset (50 video sequences). The former two methods aim at image saliency while the latter four are designed for video saliency.

\subsubsection{Qualitative Results}
Qualitative comparisons are presented in Fig. \ref{fig:fig6}, where the top line shows example video frames and the second line shows the ground truth detection results of salient objects.
As seen, the image saliency method \cite{jiang2013saliency} without deep learning, unsurprisingly, faces difficulties in dynamic scenes, due to the lack of inter-frame information and utilization of hand-crafted features.
The video saliency methods \cite{wang2015consistent,wang2015saliency} generate more visually promising results, but suffer higher computation load (which will be detailed in Sec. \ref{section5.4}) and show relatively weak performance with complex background. As for \cite{li2015visual}, it's an image saliency model but exhibits competitive performance with above bottom-up video saliency approaches, which demonstrates the power of deep learning model in saliency detection. However, we can observe the proposed algorithm captures foreground salient objects more faithfully in most test cases. In particular, the proposed algorithm yields good performance on some challenging scenarios, even for blurred backgrounds (\textit{lion01}), various object motion patterns (\textit{parkour}) or large shape deformation (\textit{soapbox}). This can be attributed to our video data synthesis, which offers diverse scene information and rich motion patterns.
Based on this, our method is able to learn both static and dynamic saliency information and detects salient moving objects accurately despite similar appearance to the background.

\subsubsection{Quantitative Results}
We report quantitative evaluation results on three widely used performance measures: precision-recall (PR) curves, F-measure and MAE.

We first employ precision-recall (PR) curves for performance evaluation. Precision corresponds to the percentage of salient pixels
correctly assigned, while recall corresponds to the fraction of detected salient pixels in relation to the ground truth number of salient pixels. For each saliency map, we vary the cutoff threshold
from 0 to 255 to generate 256 precision and recall pairs,
which are used to plot a PR curve.

The F-measure is the overall performance measurement computed by the weighted
harmonic of precision and recall:
\begin{equation}
    \begin{aligned}
    \text{F-measure} = \frac{(1+\beta^2)\times \text{precision}\times \text{recall}}{\beta^2\times \text{precision}+\text{recall}},
    \end{aligned}
\end{equation}
where we set $\beta^2 = 0.3$ to weigh precision more than recall
as suggested in \cite{Achanta2009}. For each saliency map, we derive a sequence of F-measure
values along the PR-curve with the threshold varying from 0 to 255.

\begin{figure*}%%[t]
 %%tr = 0.006, ts = 0.008
  \centering
     \includegraphics[width=\linewidth]{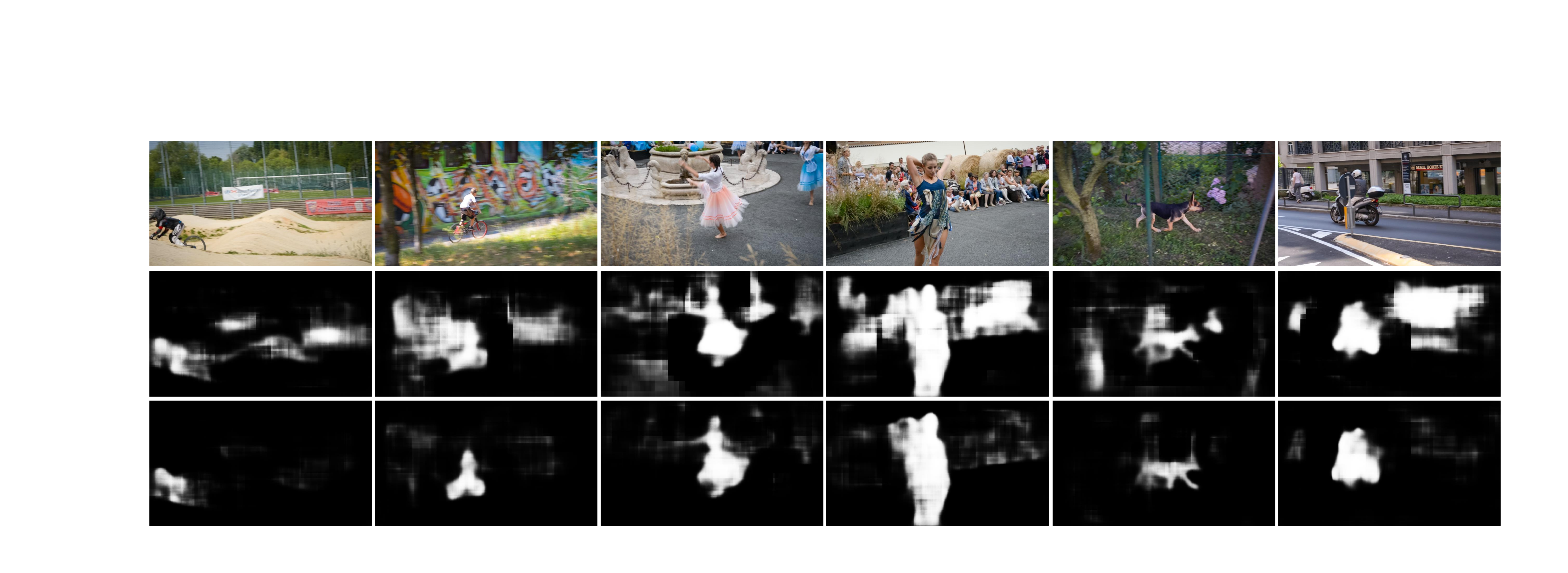}
\caption{Qualitative comparison between our static saliency results and final spatiotemporal saliency results. From top to bottom: input frame images, saliency results via our static saliency network, and spatiotemporal saliency results via our whole video saliency model.}
 \label{fig:fig8}
\end{figure*}

%%%%%%%%%%%%%%%%%%%%%%%%%%%%%%%%%%%%%%%%%%%%%%%%%%%%%%%%%%%%%%%%%%%%%%%%%
\begin{table*}%% [tbp]
\centering
\caption{Assessment of individual modules and variants of our deep saliency model on the test set of FBMS dataset \cite{brox2010object} and the DAVIS dataset \cite{Perazzi2016} using MAE. Lower values are better.}
\label{table2}
\begin{tabular}{c|c||c|c||c|c}  % {lccc}
\hline
%\multirow{1}*{} &video                   &\!\!frames\!\!
\multirow{2}*{aspect}
&\multirow{2}*{variant}
&  \multicolumn{2}{c||}{FBMS} & \multicolumn{2}{|c}{DAVIS}\\
\cline{3-6}
&& MAE(\%)       &$\Delta$\!MAE(\%)     &MAE(\%)    &$\Delta$\!MAE(\%) \\
\hline
\hline
\multirow{1}*{} &whole model   &\textbf{7.65}      &-   &\textbf{6.36}   &-  \\
\hline
\hline
\multirow{2}*{module}&Static model in Sec. \ref{section3.2}   &8.19      &+0.54   &7.17   &+0.81  \\
&Dynamic model in Sec. \ref{section3.3}   &9.43      &+1.78   &8.32   &+1.96  \\
\hline
\hline
\multirow{6}*{Training}
&Training set ~~\!\textit{i}: only using image data~($1.50\times10^5$)      &9.27      &+1.62  &7.53   &+1.17  \\
&Training set ~\textit{ii}: only using video data~($0.03\times10^5$)      &24.5      &+16.8   &23.9   &+17.5  \\
&Training set \textit{iii}: reduced training data~~~\!($1.00\times10^5$)     &9.14      &+1.48   &7.54   &+1.18  \\
&Training set \textit{iv}: reduced training data~~~\!($0.50\times10^5$)     &10.7      &+3.08   &9.13   &+2.77  \\
&Training set ~\!\textit{v}: reduced training data~~~\!($0.10\times10^5$)       &12.8      &+5.18   &10.9   &+4.58  \\
&Training set \textit{vi}: reduced training data~~~\!($0.05\times10^5$)     &13.5      &+5.83   &12.7   &+6.39  \\
\hline
\end{tabular}
\end{table*}

As neither precision nor recall considers the
true negative saliency assignments, the mean absolute error
(MAE) is also introduced as a complementary measure. MAE is defined as the average per-pixel difference
between an estimated saliency probability map $P$ and its corresponding ground truth $G$. Here,
$P$ and $G$ are normalized to the interval [0, 1]. MAE is computed as£º
\begin{equation}
    \begin{aligned}
    \text{MAE} = \frac{\sum_{i=1}^{h\times w}|P(\textbf{x}_i)-G(\textbf{x}_i)|}{h\times w},
    \end{aligned}
\end{equation}
where $h$ and $w$ refer to the height and width of the input frame image. MAE is meaningful
in evaluating the applicability of a saliency model in a
task such as object segmentation.

The precision-recall curves of all methods are reported in
Fig. \ref{fig:fig7}-a. As shown, our method significantly outperforms the
state-of-the-art both on the FBMS dataset \cite{brox2010object}, and the DAVIS dataset \cite{Perazzi2016}. Our saliency
method achieves the best precision rates, which demonstrates our saliency
maps are more precise and responsive to the actual salient information.
The F-scores are depicted in Fig. \ref{fig:fig7}-b, in which our model achieves better scores than other methods.
Similar conclusions can be drawn from the MAE. In
Fig. \ref{fig:fig7}-c, our method achieves the lowest MAE
among all compared methods.

\subsection{Validation of the Proposed Method}
\label{section5.3}
To exhibit more details of our algorithm and objectively evaluate
the contribution of different phases in the proposed saliency model, we report the evaluation
of each of the components described in Sec. \ref{section3} and different variants of the proposed saliency model.
We experiment on the test set of the FBMS dataset \cite{brox2010object}, and the DAVIS dataset \cite{Perazzi2016} and measure the performance using precision recall curve and MAE.

\subsubsection{Ablation study}
We first study the effect of each module of our deep saliency model. In Fig. \ref{fig:fig8}, we present qualitative comparison between static saliency from our static network (in Sec. \ref{section3.2}) and final spatiotemporal saliency results from our whole model (in Sec. \ref{section3.3}). It can be observed, due to the lack of dynamic information, the static saliency model faces difficulties distinguishing salient objects from clutter background in dynamic scenes. Via comprehensively utilizing static and dynamic saliency stimuli, our deep video saliency model is
able to estimate more accurate spatiotemporal saliency maps.

For quantitatively examining the performance of our static saliency network, we directly use the static saliency maps generated by the static network as final saliency estimates. From Table \ref{table2}, we can observe decreased performance (7.65$\rightarrow$8.19 on FBMS, 6.36$\rightarrow$7.17 on DAVIS), due to the lack of dynamic saliency information. Similarly, we train a dynamic network without considering static saliency as prior using the same training data. We attribute this to the difficulty of directly capturing dynamic saliency information from two successive frames without any saliency prior or extra motion information. We can draw two important conclusions. First, the fusion of static model and dynamic model
improves on both. Second, taking static saliency as prior information makes training the dynamic model easier and yield more accurate prediction.

\subsubsection{Training strategy}
We also explore the effect of different training strategies. We first study the influence of our synthetic video data generation strategy in Sec. \ref{section4}. We train our deep saliency model only using the synthetics from image data. Although the real video data occupy a small percentage of the training, we can still see a decrease in MAE (7.65$\rightarrow$9.27 on FBMS, 6.36$\rightarrow$7.53 on DAVIS) when we only use synthetic data. The small performance decrease verifies the effectiveness of our data augmentation technique; on the other hand, it suggests the synthetics should not completely replace the real video data. We further explore the performance of our model only using video data ($0.03\times10^5$ frame pairs). Unfortunately, our model suffers over-fitting due to the high similarities of scenes within same video. This also demonstrates the importance of our synthetic video data generation.

We next study the influence of the amount of training data. When we reduce the amount of training data, we can observe performance decrease. This indicates that the deep-learning model is data-driven. Or, conversely, the increase of training data will lead to improved performance.

\subsection{Runtime Analysis}
\label{section5.4}
\begin{figure}[t]
 %%tr = 0.006, ts = 0.008
  \centering
     \includegraphics[width=0.98 \linewidth]{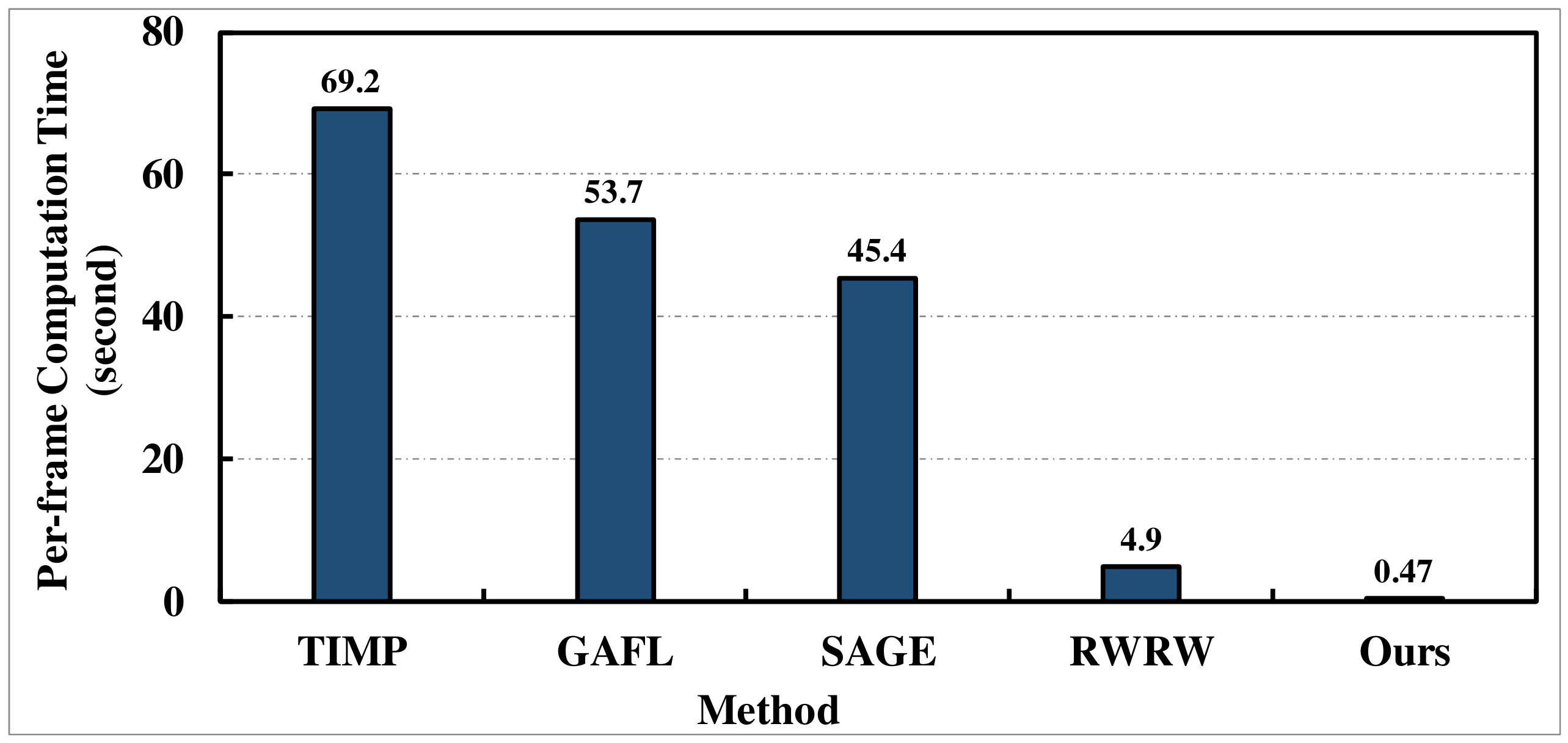}
\caption{Computational load of our method and the state-of-the-art video saliency methods for
processing a 480p video.}
 \label{fig:fig9}
\end{figure}

Here we consider the speed of our saliency method. Our computing platform includes Intel Xeon E7 CPU (12 cores)
with 64 GB memory and Nvidia Geforce TITAN X GPU. We do not count I/O time, and do not allow
processing multiple images in parallel. The time consumption, of our method compared against other video saliency
methods \cite{zhou2014time,wang2015consistent,wang2015saliency,kim2015spatiotemporal} are presented in Fig. \ref{fig:fig9}.

From Fig. \ref{fig:fig9} we can learn that, run time efficiency is the
major bottleneck for the usability of previous video saliency
algorithms, as a substantial amount of time is spent computing motion or edge information. In contrast, our method computes 480p saliency masks in as little as 0.47 seconds, which is much faster than traditional video saliency methods. Our method does not
rely on optical flow, edge maps or other pre-computed information, resulting in roughly an order of magnitude faster processing speed.

\section{Conclusion}
\label{section6}
In this work, we have presented a deep learning method for fast video saliency detection
using convolutional neural networks. The proposed deep video saliency model has two modules, namely static saliency network and dynamic saliency network, which are designed for capturing spatial and temporal statistics of dynamic scenes. The saliency estimates from the static saliency network is incorporated in the dynamic saliency network, which enables our method to automatically learn the way of fusing static saliency into dynamic saliency detection and directly produce final spatiotemporal saliency results with less computation load. Furthermore, we proposed a novel data augmentation technique for synthesizing video data from still images, which enables our deep saliency model to learn generic spatial and temporal saliency and prevents overfitting.

Experimental results on two databases, namely FBMS and DAVIS, have shown that our
proposed methods can generate high-quality salience maps. Additionally, our model waives the main computational burdens of previous video saliency models based on optical flow estimation. Our saliency model is very efficient, achieving a processing frame rate of 2fps on a GPU.

%\bibliographystyle{IEEEtran}

\vfill


\end{document}